\documentclass[conference]{IEEEtran}
\IEEEoverridecommandlockouts
\usepackage{cite}
\usepackage{amsmath,amssymb,amsfonts}
\usepackage{graphicx}
\usepackage{textcomp}
\usepackage{xcolor}
\usepackage[shortlabels]{enumitem}
\usepackage[caption=false,font=normalsize,labelfont=sf,textfont=sf]{subfig}

\usepackage{algpseudocode} 
\usepackage{booktabs}
\usepackage{multirow}
\usepackage{doi}
\usepackage{hyperref}
\usepackage{booktabs}
\usepackage{adjustbox}
\usepackage{subcaption}
\usepackage{makecell}
\usepackage{diagbox}
\usepackage{svg}
\usepackage{afterpage}
\usepackage{stfloats}
\usepackage{fontawesome}
\usepackage[a4paper,top=19mm,bottom=43mm,left=13mm,right=13mm,columnsep=4mm]{geometry}  
\usepackage{times}
\def\BibTeX{{\rm B\kern-.05em{\sc i\kern-.025em b}\kern-.08em T\kern-.1667em\lower.7ex\hbox{E}\kern-.125emX}}
\begin{document}

\title{A Novel Frequency-Spatial Domain Aware Network for Fast Thermal Prediction in 2.5D ICs}

\author{  
    \IEEEauthorblockN{Dekang Zhang\textsuperscript{1}, Dan Niu\textsuperscript{2}, Zhou Jin\textsuperscript{3}, Yichao Dong\textsuperscript{2}, Jingweijia Tan\textsuperscript{4} and Changyin Sun\textsuperscript{5}}  
    \IEEEauthorblockA{{1.}School of Integrated Circuits, Southeast University, Nanjing 210096, China}  
    \IEEEauthorblockA{{2.}School of Automation, Southeast University, Nanjing 210096, China}  
    \IEEEauthorblockA{{3.}Super Scientific Software Laboratory, China University of Petroleum-Beijing, Beijing, China}  
    \IEEEauthorblockA{{4.}College of Computer Science and Technology, Jilin University, Changchun, China}  
    \IEEEauthorblockA{{5.}School of Artificial Intelligence, Anhui University, Hefei 230601, China}   
\thanks{
This work was supported by the National Natural Science Foundation of
China (No. 62374031), Natural Science
Foundation of Jiangsu Province under Grant BK20240173, National Key R \& D Program of China (No.
2022YFB4400400), National Natural Science Foundation of China (No. 62204265, 62234010). (Corresponding authors: Dan Niu.).
}
}

\maketitle
\begin{abstract}
In the post-Moore era, 2.5D chiplet-based ICs present significant challenges in thermal management due to increased power density and thermal hotspots.
Neural network-based
thermal prediction models can perform real-time predictions for
many unseen new designs. However, existing CNN-based and GCN-based methods cannot effectively capture the global thermal features, especially for high-frequency components, hindering prediction accuracy enhancement. In this paper, we propose a novel frequency-spatial dual domain aware prediction network (FSA-Heat) for fast and high-accuracy thermal prediction in 2.5D ICs. It integrates high-to-low frequency and spatial domain encoder (FSTE) module with
frequency domain cross-scale interaction module (FCIFormer)
to achieve high-to-low frequency and global-to-local thermal dissipation
feature extraction. Additionally, a frequency-spatial hybrid loss (FSL) is designed to effectively attenuate high-frequency thermal gradient noise and spatial misalignments. The experimental results show that the performance enhancements offered by our proposed method are substantial, outperforming the newly-proposed 2.5D method, GCN+PNA, by considerable margins (over 99\% RMSE reduction, 4.23X inference time speedup). Moreover, extensive experiments demonstrate that FSA-Heat also exhibits robust generalization capabilities.

\end{abstract}


\section{Introduction}



 
To further extend Moore’s laws, 2.5D/3D chiplet-based technology is a promising technique, which integrates multiple small chips with different functionalities and manufacturing processes in a higher-density and lower-power design \cite{R1},\cite{R2},\cite{R3}. However, the large-scale and densely integrated heterogeneous chiplets simultaneously exacerbate power density and thermal hotspots \cite{R4}, compromising system-level reliability and lifespan. Thermal analysis computes on-chip temperature
distributions to check for hotspots, which
provides the optimal thermal-aware placement/floorplanning solutions
to aid quick design turnaround \cite{R5}. Therefore, accurate and fast thermal simulation, which will be performed several times, is highly desired in various thermal-aware design optimization tools.



The commercial solvers such as Celsius, ANSYS, and COMSOL widely employ the finite element method (FEM) for thermal simulation analysis \cite{R6}, which provides the best accuracy and flexibility but is computationally expensive. Current open-source thermal simulation tools further utilize the finite difference method (FDM) to simplify heat conduction equations and accelerate the thermal analysis \cite{R7}. Nonetheless, these thermal simulators provide accurate temperature estimations but consume extensive computational resources. Repeated thermal simulations during the design optimization process for complicated tasks will be unaffordable.

Recently, neural networks have been introduced to accelerate the time-consuming calculation of discrete thermal equations and perform efficient thermal analysis with high speed and accuracy. First, convolutional neural network (CNN)-based thermal prediction methods (e.g., ThermEDGe) \cite{R8}, \cite{R9}, which are mainly applied to 2D analysis, transform the thermal prediction into an image prediction problem. They demonstrate high accuracy and generalization capabilities even with the limited dataset. However, due to local perception properties and low-frequency bias \cite{R10}, CNNs often cannot perceive the long-range relationships and the varying heat dissipation gradients in each heat transfer layer, which affects the 2.5D thermal prediction accuracy owing to varying heat source size and density. In addition, graph-based and neural operator-based methods have also been proposed \cite{R11}, \cite{R12}. In \cite{R11}, the GCN architecture with graph attention network is designed to achieve 2.5D chiplet thermal prediction by aggregating information from local neighbors. However, due to ineffective global feature extraction, these CNN- and GNN-based methods encounter great challenges in higher simulation accuracy, speed, and better generalization for more fine-grained and complex 2.5D chiplet designs. Note that, widely-used global self-attention in Transformers can perceive global information but primarily focuses on the low-frequency features. High-frequency global information extraction still needs to be developed, which is vital to enhance the prediction accuracy for thermal hot spots and thermal dissipation gradients \cite{R13}. Note that, accurate thermal dissipation gradient prediction can guide high-efficiency grid partitioning for higher-accuracy simulations.

In this work, we propose a novel frequency-spatial aware prediction network named FSA-Heat, which integrates high-to-low frequency and spatial domain encoder (FSTE) module %
with frequency domain cross-scale interaction module (FCIFormer) to extract multi-frequency global-to-local thermal conduction features, achieving significantly fast and accurate thermal and dissipation gradient prediction for 2.5D-chiplet designs. The main contributions are as follows:
\begin{itemize}
  \item To the best of our knowledge, our work is the first frequency-spatial dual domain aware thermal and gradient prediction network (FSA-Heat), which captures multi-frequency global-to-local thermal dissipation features and achieves significant improvements in speed and accuracy over the newly-proposed GCN+PNA 2.5D model.
  
  

  \item In FSA-Heat, a novel high-to-low frequency and spatial domain encoder (FSTE) module is proposed, which integrates adaptive multi-frequency embeddings for high-to-low thermal dissipation gradient features and 3D CNN for detailed thermal dissipation spatial features. Furthermore, a frequency domain cross-scale interaction module (FCIFormer) is designed to leverage cross-attention for multi-scale encoder layers interaction, addressing the semantic gap fusion between layers and achieving global-to-local thermal dissipation perception in the frequency domain.

  \item A frequency-spatial hybrid loss (FSL) is proposed to effectively attenuate high-frequency thermal gradient noises and spatial structure misalignments in the 2.5D multi-layer temperature fields.

 
\end{itemize}

\begin{figure*}[t]
\renewcommand{\captionfont}{\fontsize{8}{9}\selectfont}  

\vspace{-1.0em}
\centering
\includegraphics[width=\linewidth]{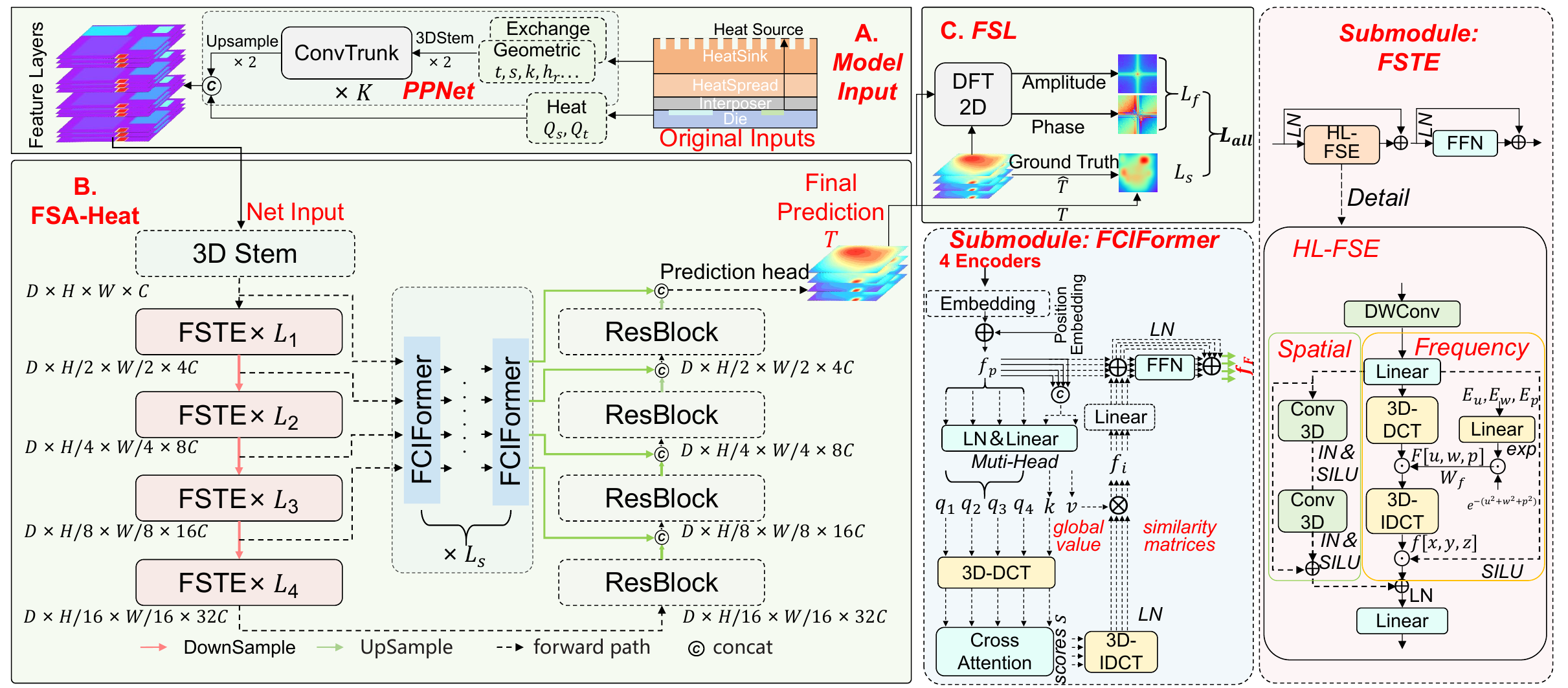}
\caption{Proposed end-to-end chiplet thermal and its gradient prediction network (FSA-Heat) and detailed structures of each module. }
%
\label{figure1}
\vspace{-1.0em}
\end{figure*}

\section{Preliminaries}


The traditional chiplet-based steady-state thermal analysis will solve the Poisson equation degenerated from the generalized heat conduction equation, as shown in Eq. (1) \cite{R14}.

\begin{equation}
k_x \frac{\partial ^2T}{\partial x^2}+k_y\frac{\partial ^2T}{\partial y^2}+k_z\frac{\partial ^2T}{\partial z^2} +Q_s=0
\label{Eq:heat_eq}
\end{equation}
where $T$ represents the target temperature, $Q_s$ denotes the power dissipation of heat source per unit volume, $(x, y, z)$ is the coordinate of the discrete grid, and $k_{n=x,y,z}$ represents the thermal conductivity in direction $\mathbf {n}$.

During the heat conduction process, the conventional finned heat sink on the top of the chiplet is employed as a convective boundary under the Neumann boundary condition $ - h{\left({\frac{{\partial T}}{{\partial \mathbf {n}}}} \right)}= \overrightarrow{q}$, where $h$ denotes the heat transfer coefficient and $\overrightarrow{q}$ is the heat flux density towards the ambient temperature. The remaining surfaces are treated as adiabatic boundaries under the Neumann boundary condition $ - k_n{\left({\frac{{\partial T}}{{\partial \mathbf {n}}}} \right)} = 0$. 

Based on the compact thermal model (CTM) \cite{R7}, the FEM is employed to solve the heat conduction equation. 
The equation is numerically discretized and modeled as a thermal resistance network $G$. 
Then, the mesh temperature can be obtained by solving the linear system of equations $GT =Q_s$. The generalized thermal conductance of each node in $G$ is $g = \sum g_{conv} + \sum g_{cond}$. Specifically, $g_{cond}$ represents the thermal conduction conductance, which is calculated by the length along the heat flux, the area perpendicular to the heat flux, and the thermal conductivity. $ g_{conv}$ is thermal convection conductance related to heat transfer coefficient, radiator convection area.

\section{Methodology}
The overall pipeline of FSA-Heat is illustrated in Fig. 1. Its Encoder part includes four encoder layers and each layer consists of $ L_i$ stacking frequency-spatial domain thermal encoder blocks (FSTE) to extract high-to-low thermal dissipation gradient features and spatial features. Moreover, the conventional skip-connections between Encoder and Decoder part are replaced by the proposed FCIFormer module, which achieves the input multi-scale encoder layers interaction and global-to-local thermal dissipation perception. The output of the FCIFormer is concatenated with the four decoder layers, which employ the residual convolution blocks. Finally, a scaled prediction head is used to generate the 2.5D thermal prediction. In FSA-Heat training, it performs backpropagation with the designed frequency-spatial hybrid loss (FSL). In Fig. 1, \( D \), \( H \), \( W \), and \( C \) represent the depth, height, width, and number of channels of the feature map, respectively.

\subsection{Model Inputs and PPNet}
In this work, the original inputs can be categorized into three groups: geometric properties (material thickness $t$, length $l$, width $w$, heatsink side length $s$), heat exchange conditions (convective resistance coefficients $h_r$, thermal conductivity $k$), and heat source parameters (layer-wise heat source power $Q_s$, total heat source power $Q_t$). $Q_t$ is established by layer-wise power allocation through $Q_t^{[i]} = \sum Q_s^{[i]}$ where $i$ is layer-wise chiplet activity profiles.
In the CTM, the geometric parameters and heat exchange conditions are first handled to obtain the thermal conduction conductance $G$ and then the thermal analysis can be simplified to solving the linear equations. Inspired by this, in this work the geometric parameter inputs and heat exchange condition inputs are also first preprocessed by a designed parameter preprocessing network (PPNet) to learn the different heat dissipation capabilities of the material layers. Note that, the time-consuming $G$-labeling for PPNet is not required and the PPNet and FSA-Heat are jointly trained to achieve meshless end-to-end thermal prediction. In PPNet, the two group inputs are first downsampled through a 3D-stem layer for dimension reduction and subsequently fed into the $K$-layer convolutional trunk to obtain fusion maps. Then, the preprocessed geometric and heat exchange parameters are obtained by the upsampling operation. Finally, the preprocessed geometric and heat exchange parameters and original heat source parameters are concatenated as the input of the FSA-Heat. Notably, the parameters between layers will not conduct convolutional interaction in order to preserve the distinctive characteristics of each layer.



\subsection{Frequency-spatial aware prediction network: FSA-Heat}
In this section, the proposed frequency-spatial domain thermal
encoder block (FSTE) works with the frequency domain cross-scale interaction module (FCIFormer) to achieve global-to-local and high-to-low frequency thermal dissipation features in the frequency and spatial dual domains, achieving higher-accuracy temperature and gradient prediction.


\subsubsection{Frequency-spatial domain thermal encoder: FSTE}


Currently, the majority of thermal prediction networks just
extract features in the spatial domain, disregarding the abundant
global information in the frequency domain. Moreover, widely-used global self-attention in Transformers tends to focus on the low-frequency features, potentially neglecting crucial high-frequency information to some extent \cite{R15}. To address these problems, FSTE module is proposed to enable concurrent learning of thermal conduction features from both spatial and frequency domains. Moreover, thermal conduction frequency features are also beneficial for accurate thermal dissipation gradient prediction. 

Similar to ViTs block \cite{R16}, the self-attention mechanism is replaced by the proposed HL-FSE  (High-to-low frequency and spatial domain embedding) block and the feed-forward network (FFN) is retained in the FSTE module, as shown in the top right corner of Fig. 1. Layernorm (LN) is used before every block, while residual connections are after every block. 

In HL-FSE, a depth-wise convolution layer is first utilized and then two branches are followed. First, the frequency domain branch is to capture the three-dimensional anisotropy frequency features in each frequency band to enable the flexible combination of both high- and low-frequency thermal dissipation gradient information. Additionally, the spatial domain branch is responsible for extracting local spatial features through a residual block consisting of 2-layer 3D convolution, instance norm (IN),
and SILU activation layers.  

In the frequency domain branch, we initially convert the mapped spatial features $f(x,y,z)$ from a linear layer into the frequency domain using 3D discrete cosine transform (3D-DCT), as depicted in Eqs. (2) and (3). Here, $F(u,w,p)$ represents the frequency domain feature maps.


\begin{multline}
F[u,w,p] = \alpha_{u,M} \alpha_{w,N} \alpha_{p,P} \sum_{x=0}^{M-1} \sum_{y=0}^{N-1} \sum_{z=0}^{P-1} f[x,y,z]\\  
\cdot C_{u,x,M} \cdot C_{w,y,N} \cdot C_{p,z,P} 
\label{Eq:DCT}
\end{multline}

\begin{equation}
 C_{u,x,M} = \cos \left[ \frac{\pi u(2x+1)}{2 \cdot M} \right]
\end{equation}
where the coefficient \(\alpha_{u,M}\) is defined as \(\alpha_{u,M} = \sqrt{\frac{2}{M}}\) for \(u \neq 0\), and \(\alpha_{u,M} = \frac{1}{\sqrt{M}}\) for \(u = 0\). And the same definitions are for $\alpha_{w,N}$, $ \alpha_{p,P} $. Moreover, $C_{w,y,N}$ and $C_{p,z,P}$ also utilize the same definition as $C_{u,x,M}$ in Eq. (3).

Then, inspired by the absolute position embeddings in ViTs \cite{R16}(despite in the frequency domain), to adaptively modulate the feature weights of different frequency bands and accurately predict the thermal dissipation gradients, $F[u,w,p]$ combines with learnable frequency weights $W_f$, as shown in Eq. (4). Different positions in the frequency domain correspond to different frequency values, which imply distinct thermal gradient patterns. By using $W_f$, HL-FSE can enhance/depress these thermal patterns within each feature channel.

\begin{equation}
W_f =  e^{-E_u\cdot u^2+E_w\cdot w^2+E_p\cdot p^2}
\end{equation}

In Eq. (4), due to the anisotropy of thermal conduction, $W_f$ is embedded from three-direction ($u,w,p$) and high-to-low frequency bands \( E_u, E_w, E_p \) that correspond to distinguish the three thermal conduction directions ($x,y,z$) in the spatial domain, respectively.

Next, by multiplying the $W_f$
and performing 3D inverse discrete cosine transformation (3D-IDCT, Eq. (5)), the frequency branch is transformed back into the spatial domain and then the multiplicative gating signal attention adjust the frequency features dynamically by weighting feature importance. 

\begin{multline}
f[x,y,z]=\alpha _{x,M}\alpha _{y,N}\alpha _{z,P}\sum_{u=0}^{M-1}{\sum_{w=0}^{N-1}{\sum_{p=0}^{P-1}{F}}}[u,w,p] \cdot W_f \\ \cdot C_{u,x,M} \cdot C_{w,y,N} \cdot C_{p,z,P}
\label{Eq:IDCT}
\end{multline}

Note that, in the frequency domain branch, the global receptive field can be obtained since each element in the frequency domain obtained by 3D-DCT incorporates information from all patches in the input feature map space. Therefore, HL-FSE can effectively capture global-to-local thermal dissipation features under different frequencies by fusing features in both spatial and frequency domains. Moreover, the training and testing efficiency is also high due to the high parallelizability of 3D-DCT and 3D-IDCT operations. Importantly,  the 3D-DCT and 3D-IDCT utilize even symmetry continuation to satisfy the implicit Neumann adiabatic boundary condition at the edges due to the even symmetry of the boundaries. The spatial branch uses reflection padding mode in its convolution operations to maintain smoothness at the boundaries and also to satisfy the adiabatic boundaries.

\subsubsection{Frequency domain cross-scale interaction module: FCIFormer}


It is known that encoder layers typically reduce the resolution of feature maps gradually, which can lead to the neglect of small-size heat sources and cause semantic loss. Moreover, the shallower encoder layer contains more raw physical parameter information, which leads to semantic gap between these features and the thermal dissipation predictions in the decoder \cite{R16}. In this work, not direct skip connections but $L_f$-layer stacking FCIFormers are proposed to further fuse the multi-scale global-to-local thermal dissipation features from the multi-scale FSTE outputs through cross-attention in the frequency domain, which can effectively alleviate the semantic gaps and achieve high-accuracy thermal prediction under largely varying heat source sizes and power densities. 


To achieve multi-scale encoder feature interaction, FCIFormer first element-wise adds the four-scale encoder embeddings with the position embedding to the embeddings $f_p$. As shown in Fig. 1, $f_p$ and their concatenated versions are passed through LayerNorm (LN) and a linear layer to obtain the query $q_1$, $q_2$, $q_3$, $q_4$, and the fused embedding key $k$, value $v$, respectively. 
In order to retain frequency information extracted from encoders, we first transform four-scale query embeddings $q_1$, $q_2$, $q_3$, $q_4$ and $k$ to the frequency domain through a 3D-DCT. The 3D-DCT converts the embedding patch dimension, embedding sequence dimension, layer depth dimension into frequency domain coordinates. Then in cross-attention, four attention scores between $q_1$, $q_2$, $q_3$, $q_4$ and $k$ are calculated respectively through dot product and softmax (in Eq. (6)), which denote semantic similarity of different scales. The four-scale design covers macroscopic (global heat diffusion), mesoscopic (regional gradient) and microscopic (heat source details) features, adapting to the diversity of heat source size and power density.
\begin{equation}
S=softmax(\frac{DCT(\mathbf{q}_i^T) \cdot DCT(\mathbf{k})}{\sqrt{d_k}})
\label{Eq:DCT}
\end{equation}
where $S$ denotes the attention scores, $q_i$ and $k$ represent the multi-head query and key matrix, respectively. $d_k$ is the dimension of the value matrix $k$.

The attention scores $S$ are used as inputs for 3D-IDCT, and $f_i$ are obtained through the weighted summation between the 3D-IDCT outputs and $v$. 
Next, $f_i$ is passed through a linear layer, followed by an element-wise addition with $f_p$. Then, the result is processed to obtain $f_F$ by a FFN to enrich the semantic content. Finally, $f_F$ is upsampled to align with the corresponding decoder to address the semantic gaps.
%

\subsubsection{Decoder}
 
In each decoder block, we upsample the feature map using trilinear interpolation, followed by a $1\times1\times1$ convolution, instance normalization (IN), and GELU activation. The upsampled feature map will concatenate with the FCIFormer output $f_F$ and then input to the next decoder layer (ResBlock). The ResBlock consists of two $3\times3\times3$ conv, instance norm, and LeakyReLU activation functions to enhance the nonlinear learning ability. In the end, the top decoder layer passes through a $1\times1\times1$ conv prediction head to obtain the final thermal prediction result $T$.






\subsection{Frequency-spatial hybrid loss (FSL)}

To achieve high-accuracy thermal and its gradient prediction, we first convert the thermal prediction images into the frequency domain function $F(u,w)$ via 2D discrete Fourier transform ($DFT_{2D}$) and then design a supervision loss function in the frequency domain, as shown in Eq. (7). \( L_f \) is the frequency loss. Its first and second term corresponds to the magnitude and phase, respectively. It allows for the frequency component comparison in both magnitude and phase through $L_1$ norm. This design effectively reduces thermal gradient prediction noise.


\begin{equation}
L_f=\left\| |F(u,w)|-|\widehat{F}(u,w)| \right\|+\beta \left\| \angle F(u,w)-\angle \widehat{F}(u,w) \right\|
\label{Eq:DCT}
\end{equation}
where \( F(u,w) \) and \( \widehat{F}(u,w) \) denote the prediction and ground truth in the frequency domain, respectively. \( \beta \) is the balance factor hyperparameter. Then the frequency-spatial hybrid loss (FSL) consists of the spatial MSE Loss $L_s$ and frequency loss $L_f$, represented as $FSL=L_s+\alpha L_f$. \( \alpha \) (here is 0.5) is the harmonic hyperparameter and $L_s$=$\| T-\widehat{T} \| _2$.

\section{Experiments and Results}
\subsection{Experiments setting}

The model is trained on a Ubuntu server equipped with a GeForce RTX 3090 GPU. The widely-used open-source software HotSpot \cite{R7} is used to generate dataset, which includes 6,000 samples with four layers: heat source layer, thermal interface material (TIM) layer, heat spreader layer, and heatsink layer. 
The heat source configurations include varying quantities (ranging from 4 to 35), sizes, positions, 
and power densities (Gaussian distribution$~$($3,{1.4}^2$)). Each layer reserves 4mm padding, and the chip layer size is 18 mm$\times$18 mm. The grid division resolution is 64$\times$64. 

\begin{figure}[tp]
\centering
\includegraphics[width=\linewidth]{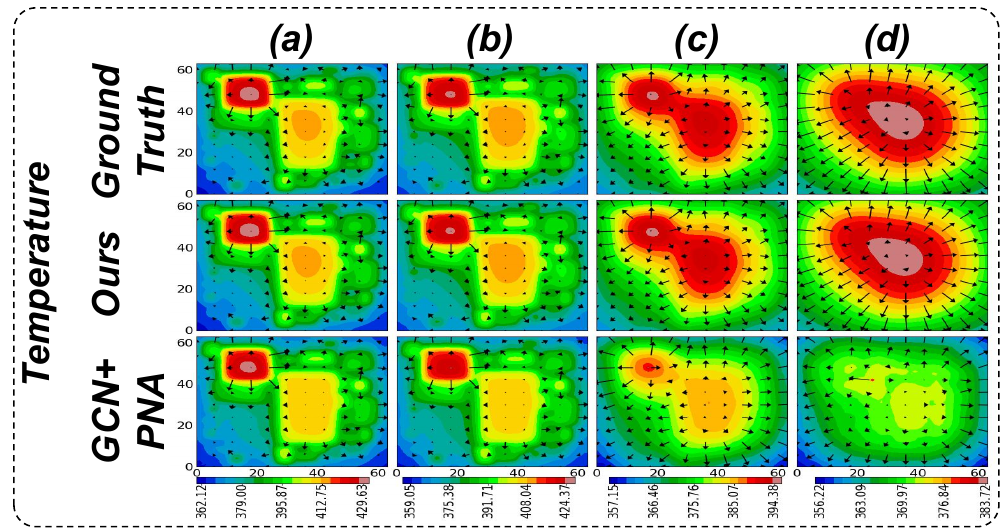}
\renewcommand{\captionfont}{\fontsize{8}{9}\selectfont} 
\vspace{-1em}
\caption{A visualization example of thermal and its gradient prediction comparisons. (a)–(d) represents the heat source, TIM, heat spreader, and heatsink layer, respectively.}
\label{figure3}
\vspace{-1.5em}
\end{figure}

\subsection{Accuracy, speed and generalization comparisons}

The proposed model will be compared with the widely-used HotSpot \cite{R7} and newly-proposed GCN-based 2.5D model \cite{R11} on the accuracy metrics such as root mean absolute error (RMSE), mean absolute percentage error (MAPE), mean absolute error (MAE), peak signal-to-noise ratio (PSNR), and the inference speed. Note that, another newly-proposed neural operator-based method DeepOHeat \cite{R12} is not compared for the following reasons: 1) it is designed for 3D chiplet; 2) the training and optimization network time on our 64 × 64 × 4 grid input (DeepOHeat is 21 × 21 × 11 grid) and our GeForce RTX 3090 is hard to afford; 3) the performance of CNN method can also be partially checked from our ablation experiment.

\begin{figure*}[hb]
\centering
\renewcommand{\captionfont}{\fontsize{8}{9}\selectfont}  
\includegraphics[width=\linewidth]{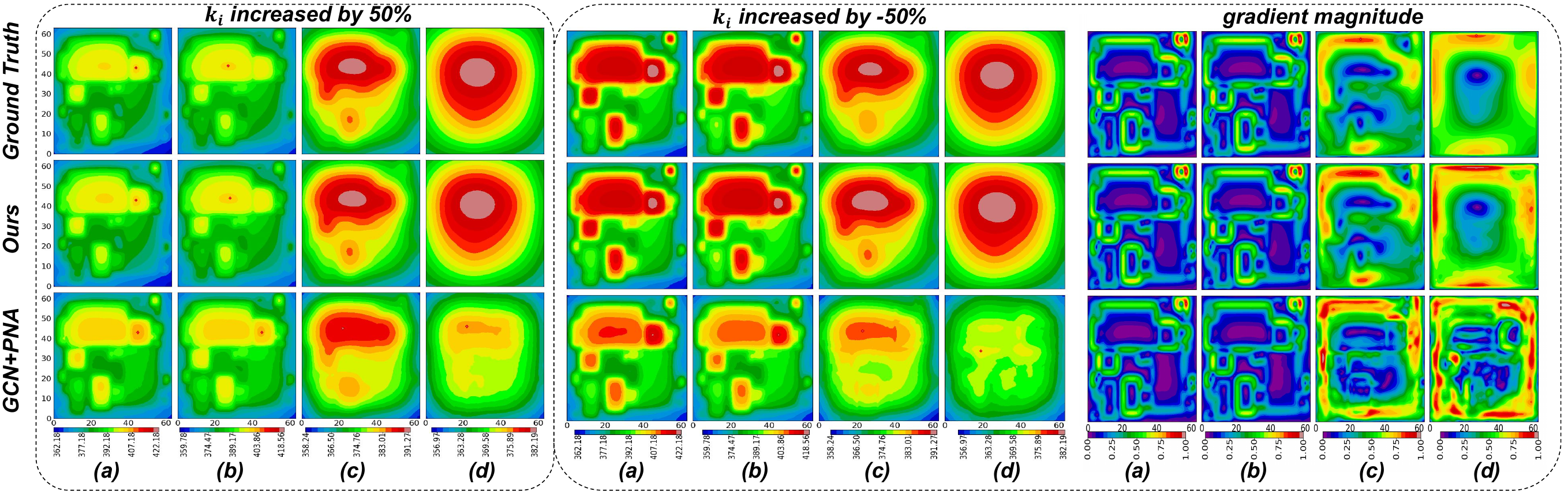}
\vspace{-2em}
\caption{Thermal and its gradient prediction comparisons between ours and GCN+PNA under unseen thermal conductivity $k_i$.}
\label{figure3}
\vspace{-1.5em}
\end{figure*}

The test results are shown in Table I,  where the proposed FSA-Heat outperforms the newly-proposed GCN+PNA \cite{R11} method with a remarkable margin in all metrics, such as RMSE (reduced over 99\%), PSNR (increased 2.37X), which means that our method achieves higher prediction accuracy and better image details.  Additionally, our model achieves the fastest inference speed with 10.58X faster than HotSpot and 4.23X faster than GCN+PNA. 
Note that we implement the GCN+PNA according to the detailed descriptions for network structure and parameters in \cite{R11}, and also achieve similar prediction accuracy at their dateset. The dataset in this work is more complicated, with much more heat sources and more diverse sizes, positions, thermal conductivities, and power densities.

\begin{figure*}[tp]
\centering
\renewcommand{\captionfont}{\fontsize{8}{9}\selectfont}  
\includegraphics[width=\linewidth]{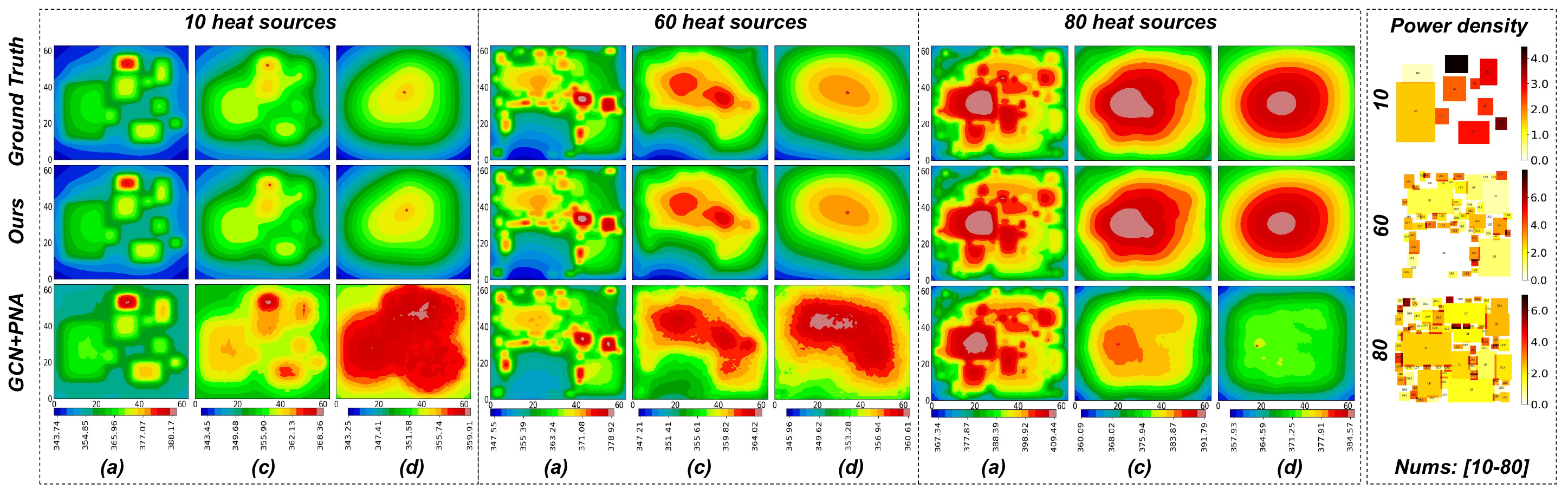}
\caption{Thermal prediction comparisons between ours and GCN+PNA under unseen varying heat source numbers.}
\label{figure3}
\vspace{-1.5em}
\end{figure*}

Moreover, the visual example maps are depicted in Fig. 2, where the temperature and its dissipation gradient predictions under four layers are given. The length and direction of arrows denote the normalized amplitude and direction of thermal gradient prediction. From this figure, all our four-layer prediction images have higher structural fidelity, with significant smaller errors
in multi-level (high and low temperature regions, especially in significantly hot spots) and multi-scale (global and local) thermal regions. Moreover, our amplitudes and directions of thermal gradients also have higher prediction accuracy, especially for large amplitude areas. By comparisons, GCN+PNA has high prediction accuracy for heat source and TIM layers, however, the prediction accuracies for heat spreader and heatsink layers decrease rapidly, especially for hot spot predictions, which is vital for thermal reliability and thermal-aware chiplet placement optimization.



In addition, the model generalization performance for the unseen testing cases is also important in thermal prediction. In this work, more difficult and unseen chiplet parameters are used to verify the model generalization performance.
\begin{table}[tp]  
    \centering  
    \renewcommand{\captionfont}{\fontsize{8}{9}\selectfont}  

    \caption{Accuracy and speed comparisons.}  
    \begin{tabular}{c|cc|c}  
        \toprule  
        Metrics & Ours & GCN+PNA & HotSpot \\ \midrule  
        $\downarrow RMSE$  & 0.0415 & 5.1424 &  -\\
        $\downarrow MAE$ & 0.0318 & 4.3484 &  -\\
        $\downarrow MAPE$ & 0.861\% & 111.235\% &  -\\
        $\uparrow PSNR$ & 68.078 & 28.659 &  -\\
        Inference Speed & 0.0549s (10.58×) & 0.2327s (2.49×) & 0.581 s \\ \bottomrule  
    \end{tabular}  
\vspace{-1em}
\end{table}


\textbf{Different thermal conductivity and heat source number:} It is clear that the thermal conductivity and heat source number are quite important for thermal dissipation. In this work, the thermal conductivities $k_i$ in each layer are given a larger change rate (50\% and -50\%) (the $k_i$ in the training dataset is from -30\% to 30\%) and the test results are shown in Fig. 3. Moreover, the heat source numbers are changed to 10, 60 and 80 (the source number in the training dataset is from 4 to 35) and the test results are depicted in Fig. 4. Note that, the two tests are conducted on 500 unseen test cases. From the Figs. 3 and 4, compared with the GCN+PNA method, the proposed FSA-Heat can also achieve higher thermal and gradient prediction accuracy for all four layers even under the unseen and large thermal conductivity and heat source number changes, which demonstrates excellent generalization performance.





\subsection{Ablation Experiments}

\begin{table}[bp]  
    \centering  
    \renewcommand{\captionfont}{\fontsize{8}{9}\selectfont}  

    \caption{Ablation experiments.}  
    \begin{tabular}{@{\hskip -1pt}c@{\hskip -4pt}c@{\hskip 3pt}c|c@{\hskip 3pt}c@{\hskip 3pt}c@{\hskip 3pt}c}  
        \toprule  
        \multicolumn{3}{c|}{Variants} & \multicolumn{4}{c}{Metrics} \\
        {FSTE} & {FCIFormer} & {FSL} & $\downarrow RMSE$ & $\downarrow MAE$ & $\downarrow MAPE$ & $\uparrow PSNR$ \\
        \midrule  
        \checkmark & \checkmark & \checkmark & 0.0415 & 0.0318 & 0.861 \%  & 68.078 \\
        \checkmark & \checkmark & spatial& {0.0532} & {0.0393} & 1.068 \%  & 65.792 \\
        \checkmark & & \checkmark & 0.0499 & 0.0383 & 0.990 \% & 66.734  \\
         w/o frequency &\checkmark & \checkmark & 0.0499 & 0.0385 & 1.047 \% & 66.384  \\
        w/o spatial &\checkmark & \checkmark & 0.0519 & 0.0398 & 1.079 \% & 66.103  \\
         spatial & & spatial & 0.0752 & 0.0534 & 1.453 \% & 62.704  \\
        \bottomrule  
    \end{tabular}
\vspace{-1em} 
\end{table}  

In this part, various ablation experiments are conducted to verify the effectiveness of the proposed FSTE, FCIFormer and
FSL modules.
First, we investigate the effectiveness of frequency-spatial dual domain thermal encoder (FSTE) by comparing the results of line 1 (proposed FSA-Heat)
with line 4 (FSA-Heat without frequency domain) and line 5 (FSA-Heat without spatial domain) in Table II. The test results demonstrate that the thermal dissipation prediction performance can be effectively enhanced
by extracting the thermal dissipation features from both the frequency and spatial domains.
Then, the effectiveness of conducting the cross-scale global-to-local thermal dissipation feature interaction in the frequency domain (FCIFormer) can be assessed by comparing line
1 with line 3. Next, by comparing line 1 and 2 (FSA-Heat without frequency loss), fusing frequency loss to conventional spatial loss (FSL) is also verified to further enhance the thermal prediction accuracy. In addition, the last line method means that just 3D CNN-based spatial domain feature extraction, skip connections between encoders and decoders, and just spatial domain loss are employed. It is a typical CNN-based method. By comparing line 1 and the last line, the effectiveness over CNN-based method can be partially verified even no direct comparison with DeepOHeat method. 


Lastly, a truth thermal example map and the prediction error maps under different ablation methods are presented in Fig. 5. It is clear that: 1) The predictions for heat spreader layer and heatsink layer are more difficult. 2) By comparing Line 1 with 2, 4, 5 of error maps, obviously, frequency domain features and loss are quite beneficial for thermal prediction, especially for heat spreader layer and heatsink layer.

\begin{figure}[tbp]
\centering
\renewcommand{\captionfont}{\fontsize{8}{9}\selectfont}  
\includegraphics[width=\linewidth]{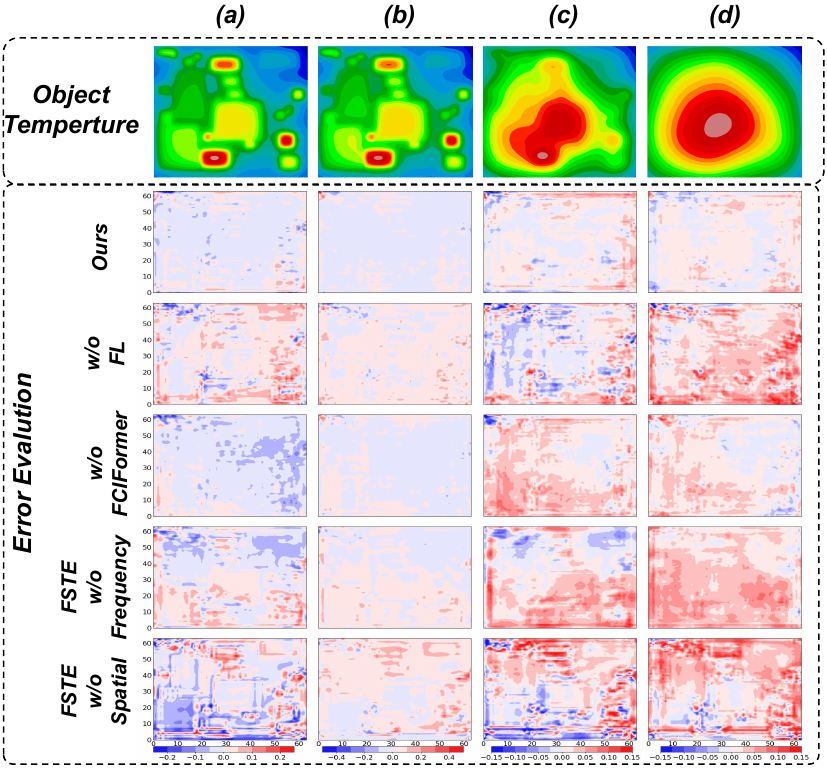}
\caption{Ablation method error visualizations.}
\label{figure3}
\vspace{-1em}
\end{figure}

\section{Conclusions}
This work presents FSA-Heat, incorporating the FSTE and FCIFormer modules for global-to-local thermal dissipation feature extraction, achieving fast and high-accuracy thermal and its gradient prediction. Moreover, a frequency-spatial hybrid loss is proposed to reduce high-frequency gradient noises. Extensive experiments reveal that FSA-Heat outperforms the newly-proposed GCN-based model in both prediction accuracy, speed, and generalization for 2.5D chiplet-based IC designs.


\ifCLASSOPTIONcaptionsoff
  \newpage
\fi

\end{document}